\documentclass[10pt,logo]{nvidiatechreport}
\providecommand{\abscontent}{}

\usepackage[authoryear,round]{natbib}

\usepackage[utf8]{inputenc} %
\usepackage[T1]{fontenc}    %

\usepackage{parskip}        %
\usepackage{url}            %
\usepackage{booktabs}       %
\usepackage{amsfonts}       %
\usepackage{nicefrac}       %
\usepackage{microtype}      %
\usepackage{xcolor}         %
\usepackage[dvipsnames]{xcolor} %
\usepackage{graphicx}
\usepackage{animate}        %
\usepackage{subcaption}
\usepackage{tabularx}
\usepackage{makecell}
\usepackage{adjustbox}
\usepackage{setspace}
\newcolumntype{M}[1]{>{\centering\arraybackslash}m{#1}}
\usepackage{float}
\usepackage{tikz}
\usetikzlibrary{positioning,shapes,arrows}
\usepackage{amsmath,amsfonts,bm, bbm,leftindex}
\usepackage{multirow}
\usepackage{comment}
\usepackage{gensymb}
\usepackage{lipsum}
\usetikzlibrary{arrows.meta, positioning, fit}
\usepackage[para]{threeparttable}
\usepackage{tikz}
\usetikzlibrary{tikzmark}
\usepackage{hyperref}
\usepackage[most]{tcolorbox}
\usepackage{fancyvrb}
\usepackage{fvextra}
\usepackage{dashrule}
\usepackage{wrapfig}

\titleformat{\paragraph}[runin]{\normalfont\normalsize\bfseries}{\theparagraph}{1em}{#1}
\titlespacing*{\paragraph}{0pt}{1ex plus1pt minus1pt}{0.5em}
\titleformat{name=\section,numberless}{\Large\bfseries\headingfont}{}{0pt}{#1}

\usepackage{amsmath,amsfonts,bm}

\def\eqref#1{equation~\ref{#1}}

\def\1{\bm{1}}

\DeclareMathAlphabet{\mathsfit}{\encodingdefault}{\sfdefault}{m}{sl}
\SetMathAlphabet{\mathsfit}{bold}{\encodingdefault}{\sfdefault}{bx}{n}

\usepackage{amssymb}
\usepackage{array}
\usepackage{colortbl}
\usepackage{enumitem}
\graphicspath{{assets/}}
\newcolumntype{L}[1]{>{\raggedright\arraybackslash}p{#1}}
\usepackage{xcolor}

\usepackage{placeins}
\newsavebox{\liberobox}
\newsavebox{\liberocap}
\makeatletter
\newcommand\@FloatFlushLoop{\par\begingroup\let\@elt\relax
  \edef\@tempa{\@deferlist\@dbldeferlist}%
  \ifx\@tempa\@empty\else\null\newpage\@FloatFlushLoop\fi\endgroup}
\newcommand\FloatFlush{\setcounter{topnumber}{9}\setcounter{totalnumber}{9}\@FloatFlushLoop
  \setcounter{topnumber}{3}\setcounter{totalnumber}{4}}
\makeatother
\newif\iftightlayout \tightlayouttrue
\iftightlayout
\AtBeginDocument{\setlength{\abovedisplayskip}{4pt plus 2pt minus 2pt}\setlength{\belowdisplayskip}{4pt plus 2pt minus 2pt}\setlength{\abovedisplayshortskip}{2pt}\setlength{\belowdisplayshortskip}{2pt}}
\newcommand{\lsn}{\looseness=-1}
\else
\newcommand{\lsn}{\looseness=-1}
\fi
\newcommand{\ourmethod}{TacitVLA}
\newcommand{\ourmethodit}{\textit{\ourmethod{}}}

\newcommand{\Tact}{T_{\text{action}}}
\newcommand{\Treason}{T_{\text{reason}}}
\newcommand{\Texec}{T_{\text{exec}}}
\definecolor{oursgreen}{HTML}{76B900}
\colorlet{oursbg}{oursgreen!12}
\newcommand{\ours}{\rowcolor{oursbg}}
\definecolor{explicitink}{HTML}{2F5FA8}
\definecolor{tacitink}{HTML}{5E9400}
\definecolor{noneink}{HTML}{8C8C8C}
\newcommand{\noreasonc}{\textcolor{noneink}{No Reasoning}}
\newcommand{\explicitc}{\textcolor{explicitink}{Explicit Reasoning}}
\newcommand{\ourmethodc}{\textcolor{tacitink}{\ourmethod{}}}

\title{\ourmethod{}: Reasoning Vision-Language-Action Models Without Reasoning Latency}

\author[1,2,$\dagger$]{Jack Lu}
\author[1]{Yurong You}
\author[1,3]{Boyi Li}
\author[1]{Yulong Cao}
\author[1]{Thomas Tian}
\author[1,4]{Chaowei Xiao}
\author[1]{Xiangyu Chen}
\author[1]{Wenjie Luo}
\author[1]{Boris Ivanovic}
\author[1,5,$*$]{Marco Pavone}
\author[1,$*$]{Yan Wang}
\affil[1]{NVIDIA}
\affil[2]{New York University}
\affil[3]{University of California, Berkeley}
\affil[4]{Johns Hopkins University}
\affil[5]{Stanford University}
\affil[$\dagger$]{Work done during an internship at NVIDIA}
\affil[$*$]{Equal advising}

\begin{abstract}
Textual reasoning before acting has improved vision-language-action (VLA) models, but at the cost of latency: every action conditioned on the reasoning text must wait for autoregressive decoding. We show that actions do not need to wait for or condition on the generated reasoning to benefit from it, because the reasoning supervision already improves the backbone representation. We introduce \ourmethodit{}, a reasoning VLA whose action expert never attends to the reasoning it generates for the current call. On long-horizon tasks where the reasoning text also serves as episodic memory, it can generate each entry during action execution rather than before acting. We evaluate it in a controlled study on four benchmarks spanning manipulation and autonomous driving, memory and non-memory tasks: RoboMME, RMBench, LIBERO, and PhysicalAI-AV. On all four, \ourmethodit{} matches the performance of its explicit-reasoning counterpart at up to 3.4$\times$ lower latency, and matches the latency of the same architecture trained without reasoning at up to +17.1\% success. In absolute terms, it surpasses the best previously published results on the RoboMME Counting suite (82.7\% vs.\ 66.8\%) and RMBench (77.5\% vs.\ 56.5\%), and reaches 72.9\% on LIBERO-Plus and a 52.5\% scene score on the public split of the PhysicalAI-AV closed-loop driving challenge. Project page: \url{https://tacitvla.github.io}.
\end{abstract}

\begin{document}

\maketitle

\abscontent

\begin{figure}[!b]
 \centering
 \includegraphics[width=0.97\linewidth]{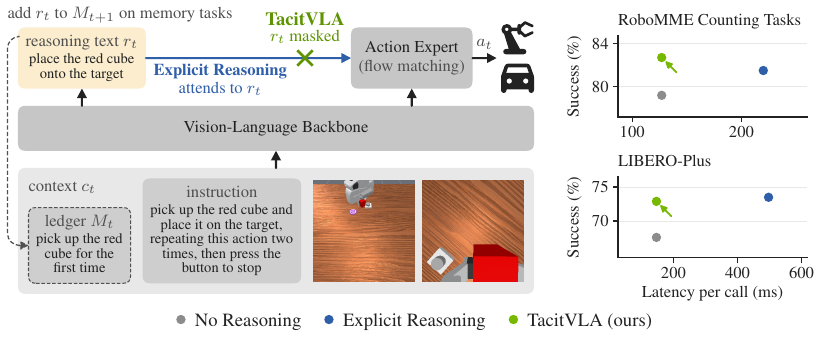}
 \caption{\lsn \textbf{\ourmethod{} retains the benefit of reasoning supervision without conditioning its actions on the generated text.} Left: how the policies act in one call. Given the context $c_t$, on memory tasks including the ledger $M_t$ of past subtasks, the backbone generates the reasoning $r_t$ and conditions the action expert that outputs the action chunk $a_t$. In \explicitc{}, the action expert attends to $r_t$, so every action chunk waits for it. \ourmethodc{} trains on the same sequence with $r_t$ masked from the action expert, bypassing the need to generate $r_t$ on non-memory tasks. On memory tasks, $r_t$ still becomes the next ledger entry (dashed loop), but is generated asynchronously during execution instead of making the action expert wait. \noreasonc{} is the same architecture trained without reasoning supervision. Right: \ourmethod{} matches Explicit Reasoning's success at 1.7$\times$ and 3.4$\times$ lower latency, and matches No Reasoning's latency at +3.5\% and +5.3\% success (Tables~\ref{tab:memory}, \ref{tab:libero}, \ref{tab:stages}).}
 \label{fig:overview}
\end{figure}

\section{Introduction}
\label{sec:intro}

\lsn Vision-language-action (VLA) models that generate a scene description, a plan, or the next subtask before acting have outperformed models that map observations directly to actions \citep{ecot2024,pi052025,deepthink2025,alpamayo2025,wla2026}. The generated text plays three roles in such a policy. (i) During training, it is a target, and the supervision shapes the backbone representation on which the action expert conditions. (ii) On long-horizon tasks, it can serve as episodic memory: each generated subtask is appended to the context, recording what the current observation no longer shows, such as how many button presses have been made. (iii) At deployment, it is an input: the action expert conditions on the reasoning text generated for the current call, and because the reasoning text is decoded autoregressively, the action must wait for that decoding to finish. For the reasoning VLAs we train on LIBERO and PhysicalAI-AV, generating the reasoning text accounts for 53\% to 70\% of the latency per call (Section~\ref{sec:latency}). 

\lsn To address the latency problem, dual-system designs run the vision-language model at a lower rate than a separate controller \citep{hirt2024,thinkact2025,helix2025}, which either does not eliminate the wait or makes the controller act on what that model produced for an earlier observation. Emerging work trains reasoning VLAs on reasoning text and deploys them without generating it \citep{zr02026,ervla2026}, yet does not answer whether that comes at a cost to performance and does not consider tasks where the generated text must be retained as memory. We therefore ask whether role (iii) is truly necessary:
\begin{center}
\emph{Must the action condition on the current reasoning, or is learning to generate it enough?}
\end{center}
\lsn If the benefit of reasoning mainly comes through role (i), generating the reasoning text adds latency without improving performance. Where the reasoning text also serves as memory (ii), it must still be generated for later calls, but this can happen during action execution rather than before (Figure~\ref{fig:overview}).

\lsn Our controlled study on four benchmarks shows that learning to generate the reasoning is enough. On each, we train three policies of the architecture recent reasoning VLAs adopt \citep{pi052025,alpamayo2025,wla2026,onetwo2025,zr02026,ervla2026}: a vision-language backbone that generates the reasoning and feeds a flow-matching action expert. \emph{Explicit Reasoning} deploys as embodied chain-of-thought does \citep{ecot2024}: its action expert attends to the reasoning it has just generated, making every action chunk wait. We introduce \ourmethodit{}, which is trained on the same sequence and differs only in masking the action expert's attention to the reasoning tokens (Figure~\ref{fig:method}), so its actions do not wait for the reasoning text. \emph{No Reasoning} is the same architecture trained without reasoning supervision, which matches \ourmethodit{}'s latency and quantifies the benefit of that supervision. The benchmarks split by how the reasoning text is used: on the \emph{memory tasks}, RoboMME and RMBench, each generated subtask becomes the next entry of a ledger that every later call conditions on. On the \emph{non-memory tasks}, LIBERO and PhysicalAI-AV, the reasoning text serves the current action only and is discarded. Our contributions are:
\begin{itemize}[leftmargin=*,itemsep=1pt,topsep=2pt]
 \item \lsn \textbf{Controlled study: an action expert that never conditions on the current reasoning matches one that does.} We train Explicit Reasoning and \ourmethodit{} on the same sequences and vary only the action expert's attention to the reasoning tokens. We evaluate on four benchmarks spanning manipulation and autonomous driving, memory and non-memory tasks. On all four, \ourmethodit{} matches the performance of Explicit Reasoning, confirming that learning to generate the reasoning is enough.
 \item \lsn \textbf{\ourmethod{}: Explicit Reasoning's success at No Reasoning's latency.} \ourmethodit{}'s latency is up to 1.7$\times$ lower than Explicit Reasoning's on the memory tasks and 3.4$\times$ lower on the non-memory tasks (Section~\ref{sec:latency}). While matching No Reasoning's latency, it improves success by up to +17.1\% (Section~\ref{sec:performance}). In absolute terms, it surpasses the best published results on RoboMME (48.1\% vs.\ 44.5\%; \citealp{robomme2026}) and RMBench (77.5\% vs.\ 56.5\%; \citealp{wla2026}), and reaches 72.9\% on LIBERO-Plus and a 52.5\% scene score on PhysicalAI-AV (Section~\ref{sec:performance}). 
\end{itemize}

\section{Related work}
\label{sec:related}

\paragraph{Reasoning before acting and its cost.} \lsn Embodied chain-of-thought trains a VLA to describe its scene, plan, and next subtask before acting and shows that the reasoning improves generalization \citep{ecot2024}. The pattern now runs through manipulation, driving, and hierarchical policies \citep{onetwo2025,wla2026,alpamayo2025,pi052025,vlaos2025}. The cost is autoregressive generation on the action path: embodied chain-of-thought policies run at 0.3 to 0.5\,Hz compared with 3 to 4\,Hz for the same architecture without test-time reasoning on the same GPU \citep{ecotlite2025}, and hybrid action decoding narrows the gap to 1.4$\times$ \citep{deepthink2025}. Other methods that reduce this latency include the reuse of a potentially stale chain across calls \citep{fastecot2025}, latent reasoning in place of tokens \citep{fastthinkact2026,laravla2026,lcdrive2025}, and, for VLA inference in general, early exit, token caching, and state-space backbones \citep{deer2024,vlacache2025,robomamba2024}. Instead of reducing the latency cost of reasoning, \ourmethodit{} removes the need for reasoning generation before action prediction while keeping reasoning as a training target, eliminating the reasoning latency.

\paragraph{Reasoning as a training target.} \lsn Several prior works train on reasoning text and deploy without generating it: implicit planning in VLA-OS \citep{vlaos2025}, $\pi_{0.5}$ deployed without subtask inference \citep{pi052025}, the reasoning pre-training and dropout recipes of ECoT-Lite \citep{ecotlite2025}, ZR-0 \citep{zr02026}, and ERVLA \citep{ervla2026}. They establish that reasoning supervision can help a policy that generates no text at deployment. However, they either do not compare with an explicit-reasoning counterpart \citep{zr02026}, or the comparison also varies how actions are decoded \citep{ervla2026} or the training mixture \citep{ecotlite2025}, switches generation off on a checkpoint trained to condition on it \citep{pi052025}, or uses a small policy trained from scratch \citep{vlaos2025}. We hold the sequence, data, prompts, and optimizer fixed, vary the action expert's attention alone, extend the comparison to memory tasks, where the reasoning text must still be generated, and measure latency beside success. Fast-WAM \citep{fastwam2026} asks the analogous question for future video and suggests that video prediction helps mainly by improving representations during training rather than by generating the future at test time. In language modeling, rationales used only as training targets improve answers \citep{hsieh2023distilling}, and chains of thought can be internalized \citep{deng2023implicit,hao2024coconut,kim2026hyperthink}.

\paragraph{Memory and asynchronous reasoning.} \lsn Memory benchmarks \citep{robomme2026,rmbench2026} require information that the current observation no longer shows. Some policies keep it as text: WLA's subtask ledger, which its action waits for \citep{wla2026}, and OneTwoVLA's latest reasoning \citep{onetwo2025}. Memory can also be represented as latent tokens \citep{memoryvla2026} or recent frames plus compressed gist tokens \citep{memorywam2026}. Hierarchical and dual-system designs run the vision-language model below the control rate, with a fast controller conditioning on what the slow model has produced \citep{hirt2024,fastinslow2025,fastecot2025}. Real-time chunking keeps actions continuous while the next action chunk is computed \citep{rtc2025}. These designs change the schedule or the memory format. \ourmethodit{} keeps one backbone feeding one action expert, holds the ledger protocol fixed, and moves the generation of each entry into the execution window to eliminate its latency at deployment.

\section{Preliminaries}
\label{sec:prelim}

\subsection{Reasoning VLAs}
\label{sec:setup}

\begin{figure}[t]
 \centering
 \includegraphics[width=\linewidth]{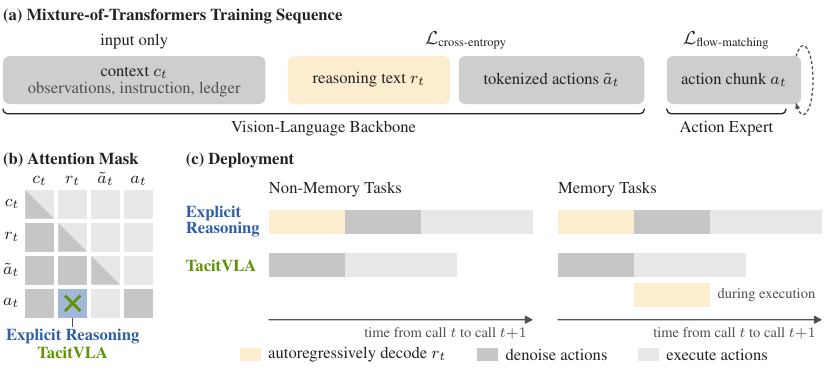}
 \caption{\lsn \textbf{Training and deployment of \explicitc{} and \ourmethodc{}.} The two policies share the training sequence and differ in one attention-mask cell, which decouples \ourmethod{}'s actions from text generation at deployment. (a) The backbone predicts the reasoning $r_t$ and the tokenized actions $\tilde{a}_t$ (an auxiliary target) under cross-entropy, and the action expert denoises the action chunk $a_t$ under flow matching. (b) Rows attend to columns. The backbone is causal. The action expert attends to $c_t$ and its own action chunk, to $r_t$ only in Explicit Reasoning, and never to $\tilde{a}_t$. The crossed cell is that difference. (c) On non-memory tasks, Explicit Reasoning needs to generate $r_t$ before denoising and executing the action chunk, while \ourmethod{} does not. On memory tasks, both generate the ledger entry $r_t$: Explicit Reasoning before action execution, \ourmethod{} during it concurrently.}
 \label{fig:method}
\end{figure}

\lsn A reasoning VLA is trained to generate reasoning text before acting \citep{ecot2024,pi052025,alpamayo2025,wla2026}. We study the form adopted by recent reasoning VLAs: a vision-language backbone generates the reasoning text and an action expert conditions on the backbone's representation \citep{pi052025,alpamayo2025,wla2026,onetwo2025,zr02026,ervla2026}. A \emph{call} of the policy maps the context to one action chunk, part or all of which is executed before the next call. The context $c_t$ at call $t$ holds the instruction and the observation: camera views, robot or vehicle state, and the history the policy keeps (Appendix~\ref{app:protocol}). On memory tasks, it also holds a ledger $M_t$ of past subtasks. From it, the backbone generates the reasoning text $r_t$, which is the next subtask, a scene description and plan, or a chain of causation, depending on the benchmark (Appendices~\ref{app:targets} and \ref{app:examples}).

\lsn The action expert generates an action chunk $a_t$ \citep{zhao2023act} by denoising \citep{diffusionpolicy2023}, with flow matching in $K$ Euler steps \citep{pi02024}. The backbone and the action expert form a Mixture-of-Transformers \citep{mot2024}: they keep separate weights but share one attention, in which the action positions attend to the backbone's keys and values layer by layer. Both train jointly on the same examples,
\begin{equation}
 \mathcal{L} \;=\; \mathcal{L}_{\text{flow-matching}}(a_t) \;+\; \lambda\,\mathcal{L}_{\text{cross-entropy}}(r_t, \tilde{a}_t),
 \label{eq:loss}
\end{equation}
\lsn where the flow-matching loss is the action expert's denoising objective on the noised action chunk \citep{pi02024}, and the cross-entropy is the backbone's next-token loss on the reasoning text and on $\tilde{a}_t$. Here, $\tilde{a}_t$ is the same action chunk in tokenized form, which the backbone predicts after $r_t$: FAST+ tokens \citep{fast2025}, or trajectory tokens in driving \citep{alpamayo2025}. Training the backbone on tokenized actions alongside a flow-matching action expert that alone produces the deployed action follows $\pi_{0.5}$, and the tokenized actions speed up training \citep{fast2025,pi052025}.

\lsn At deployment, such a policy can generate $r_t$ and then denoise the action from the backbone's representation over $c_t$ and $r_t$, so every action waits for the reasoning text. This follows the deployment pattern of embodied chain-of-thought \citep{ecot2024}, and we call it \emph{Explicit Reasoning}. The same architecture trained without reasoning supervision generates no text, and we call it \emph{No Reasoning}.

\subsection{Memory and non-memory tasks}
\label{sec:lifetimes}

\lsn The reasoning text is used in two ways. On \emph{memory tasks}, it serves the immediate action and the calls after it: the generated subtask $r_t$ is appended to the ledger, so $M_{t+1}$ is $M_t$ followed by $r_t$, and every later call conditions on it. The ledger is a symbolic memory of what the current observation does not show, and an entry names the subtask the action chunk attempted. On RMBench, we follow the form of WLA's subtask ledger \citep{wla2026}, and on RoboMME, the entries are the benchmark's own simple subgoal annotations \citep{robomme2026}. On \emph{non-memory tasks}, the reasoning text serves the immediate action only: $r_t$ describes the current scene and the action to take, and later calls do not condition on it, as in the LIBERO annotations of DeepThinkVLA \citep{deepthink2025} and the chain of causation in driving \citep{alpamayo2025}.

\subsection{Latency of a call}
\label{sec:timing}

\lsn Three stage times describe one call. $\Tact$ is the action path: visual encoding, prompt prefill, and denoising, the same three stages for every policy we train. $\Treason$ is the cost of explicit reasoning: generating $r_t$ and, where the action expert conditions on it, the backbone's pass over the generated tokens. $\Texec$ is the execution window: the physical time the committed actions take to execute before the next call is due, as in real-world deployment (Appendix~\ref{app:timing}). The \emph{latency} of a call is the time from the observation until the action chunk is ready. For No Reasoning, it is $\Tact$, whereas for Explicit Reasoning, it is $\Tact + \Treason$ on memory and non-memory tasks alike.

\section{\ourmethod{}}
\label{sec:method}

\subsection{Masking reasoning from the action expert}
\label{sec:policies}

\lsn The backbone's input for one training example is the context $c_t$ followed by the reasoning target $r_t$ and the tokenized actions $\tilde{a}_t$ (Section~\ref{sec:setup}), with the action chunk's placeholder positions appended for the action expert (Figure~\ref{fig:method}). To make a controlled comparison, Explicit Reasoning and \ourmethodit{} share this sequence, the loss in Equation~\ref{eq:loss}, the prompts, the data, and the optimizer. They differ only in the attention pattern of the action positions:
\begin{alignat}{2}
 \text{Explicit:}\quad & \pi\bigl(a_t \mid c_t, r_t\bigr),\hspace{3em} &&
 \text{the action expert attends to } c_t \text{ and } r_t; \label{eq:explicit}\\
 \text{\ourmethod{}:}\quad & \pi\bigl(a_t \mid c_t\bigr), &&
 \text{the action expert attends to } c_t \text{ only.} \label{eq:tacit}
\end{alignat}
Reasoning can thus reach the action through two pathways: the cross-entropy on the reasoning text trains the backbone whose representation the action expert conditions on, and the action expert can attend to $r_t$ itself. Explicit Reasoning has both, \ourmethodit{} only the first, and No Reasoning neither, so comparing \ourmethodit{} with No Reasoning isolates the first pathway and comparing it with Explicit Reasoning isolates the second. At deployment, Explicit Reasoning's $r_t$ is generated from $c_t$ alone, so the second pathway adds computation, not information. On the tasks we study, \ourmethodit{} matches Explicit Reasoning (Section~\ref{sec:performance}), so the gain from reasoning does not need the extra computation. Both patterns are masks over one shared sequence, so the sequence shape is identical, and \ourmethodit{} uses slightly less training compute per update because its action positions do not attend to $r_t$.

\subsection{Deferring text generation}
\label{sec:defer}

\lsn On non-memory tasks, nothing at deployment consumes $r_t$, so \ourmethodit{} does not generate it. Its latency is $\Tact$, the same as No Reasoning's. On memory tasks, \ourmethodit{} computes the action first and generates the ledger entry afterwards. At call $t$, the action expert acts from $c_t$, which already contains $M_t$. While the robot executes this action chunk, the backbone generates $r_t$, which enters the ledger for call $t+1$ exactly as under Explicit Reasoning. Generation delays the next call only if it takes longer than the execution window, and then only by the difference. \ourmethodit{}'s latency on memory tasks is therefore
\begin{equation}
 \Tact + \max(0,\, \Treason - \Texec).
 \label{eq:cost}
\end{equation}
\lsn Section~\ref{sec:latency} reports the measured $\Tact$ and $\Treason$ for No Reasoning, Explicit Reasoning, and \ourmethodit{} on all benchmarks.

\section{Experimental setup}
\label{sec:expsetup}

\paragraph{Benchmarks.}\label{sec:benchmarks} \lsn We run the comparison between No Reasoning, Explicit Reasoning, and \ourmethodit{} on four benchmarks, two memory and two non-memory (Table~\ref{tab:hosts}). On each, the three policies differ only in the attention mask, the reasoning supervision, and, on memory tasks, the ledger. On the memory tasks, No Reasoning thus measures the joint value of reasoning supervision and the ledger.

\begin{table}[t]
 \centering
 \small
 \caption{\lsn \textbf{The four benchmarks.} Each row is one architecture, trained three times on the same data: as No Reasoning, Explicit Reasoning, and \ourmethod{}. ``Reasoning text'' is what the backbone generates (Appendix~\ref{app:protocol}). Size is \ourmethod{}'s total parameter count, action expert included; on the non-memory tasks, Explicit Reasoning needs 0.05B more to generate text (Table~\ref{tab:params}).}
 \label{tab:hosts}
 \setlength{\tabcolsep}{4pt}
 \begin{tabular}{@{}llllc@{}}
  \toprule
  Benchmark & Task type & Reasoning text & Backbone & Size \\
  \midrule
  RoboMME & memory & next subtask & PaliGemma ($\pi_{0.5}$) & 3.7B \\
  RMBench & memory & next subtask & RynnBrain-2B & 2.6B \\
  LIBERO (+Plus) & non-memory & scene description and plan & RynnBrain-2B & 2.5B \\
  PhysicalAI-AV & non-memory & chain of causation & Qwen3-VL-2B & 3.9B \\
  \bottomrule
 \end{tabular}
\end{table}

\paragraph{RoboMME.} \lsn RoboMME is a 16-task memory benchmark \citep{robomme2026}. Its four counting tasks require counting repeated actions or events. Its authors find perceptual and symbolic memory complementary and suggest combining them, which we do in one backbone. Every policy carries the benchmark's perceptual memory, FrameSamp with its modulator, on all 16 tasks. The reasoning policies add the ledger on the counting tasks, trained on the benchmark's simple subgoal annotations (Appendix~\ref{app:protocol}). We train policies on the four counting tasks and on all 16 tasks.

\paragraph{RMBench.} \lsn RMBench \citep{rmbench2026} tests memory over long manipulation episodes. We use the four of its nine tasks that WLA-0 reports, 100 cases each: Blocks Ranking Try, Battery Try, Press Button, and Cover Blocks.

\paragraph{LIBERO.} \lsn LIBERO is the standard manipulation benchmark of four 10-task suites \citep{libero2023}, for which we use DeepThinkVLA's reasoning annotations \citep{deepthink2025}. To test robustness out of distribution, we also evaluate LIBERO-Plus \citep{liberoplus2025}, which applies seven perturbation axes zero-shot over 10{,}030 cases.

\paragraph{PhysicalAI-AV.} \lsn We train on a proprietary corpus of 3.8 million driving events with chain-of-causation annotations in the style of Alpamayo-R1 \citep{alpamayo2025}. The policy predicts a 6.4\,s trajectory. We evaluate in open loop on a curated set of 1{,}434 held-out segments of the public PhysicalAI-AV release \citep{physicalaiav2025} and in closed loop in AlpaSim \citep{alpasim2025}, on the 441-scene public split of its driving challenge (\citealp{alpasimchallenge2026}; Appendix~\ref{app:protocol}).

\paragraph{Training and evaluation.} \lsn On the three robot benchmarks, we train only on each benchmark's released demonstrations and compare with published results of other policies trained on them. Our policies are evaluated under each benchmark's official protocol and cases, and their success values are means $\pm$ standard deviation over three evaluation seeds. Closed-loop driving values are means over three rollouts per scene. All training runs on nodes of eight H100 GPUs, and all three policies run the same inference implementation, timed on one H100 (Appendix~\ref{app:timing}). Initializations, architectures, training settings, prompts, decoding, the ledger protocol, and the source of every published row are in Appendix~\ref{app:protocol}.

\section{Results}
\label{sec:results}

\subsection{Performance}
\label{sec:performance}

\begin{wraptable}[11]{r}{0.38\textwidth}
 \vspace{\dimexpr-1.14\baselineskip-14pt\relax}
 \centering
 \caption{\lsn \textbf{RoboMME Counting.} Success (\%) on the benchmark's four counting tasks (200 cases) of policies trained on them alone, and the benchmark's human reference \citep{robomme2026}. Shaded rows are ours, mean $\pm$ standard deviation over three evaluation seeds.}
 \label{tab:counting}
 \vspace{1.9pt}
 \small
 \setlength{\tabcolsep}{4pt}
 \begin{tabular}{@{}lc@{}}
  \toprule
  Policy & Success \\
  \midrule
  Human (oracle planner) & 88.5 \\
  \ours No Reasoning & 79.0\,{\scriptsize$\pm$\,1.0} \\
  \ours Explicit Reasoning & 86.7\,{\scriptsize$\pm$\,2.0} \\
  \ours \ourmethod{} & \textbf{89.5}\,{\scriptsize$\pm$\,1.3} \\
  \bottomrule
 \end{tabular}
 \vspace{-0.6\baselineskip}
\end{wraptable}
\paragraph{RoboMME.}\label{sec:persistent} \lsn Trained on the Counting suite alone and evaluated on its four tasks (200 cases, Table~\ref{tab:counting}), \ourmethodit{} reaches 89.5 and Explicit Reasoning 86.7, compared with 79.0 for No Reasoning. \ourmethodit{}'s 89.5 is above even the benchmark's human reference of 88.5 \citep{robomme2026}. In that reference, people only chose each high-level action and an oracle planner executed it without error, whereas \ourmethodit{} must also execute every motion. Reasoning supervision adds 10.5 points for \ourmethodit{} and 7.7 for Explicit Reasoning, so an entry generated during action execution retains the full gain of one generated before the action. \ourmethodit{}'s ledger is also as accurate: the final ledger matches the instructed count within two points of Explicit Reasoning's on each counting task in Table~\ref{tab:ledger}.

\begin{table}[t]
 \centering
 \caption{\lsn \textbf{RoboMME.} Success (\%) of policies trained on all 16 tasks, on each suite (200 cases) and on all tasks (800 cases). Shaded rows are ours, mean $\pm$ standard deviation over three evaluation seeds. Only the counting tasks have reasoning supervision. Published rows are from the RoboMME paper \citep{robomme2026}, except WeaveLA and AGM, which report their own (Appendix~\ref{app:protocol}).}
 \label{tab:memory}
 \small
 \setlength{\tabcolsep}{7pt}
 \begin{tabular}{@{}lccccc@{}}
  \toprule
  Policy & Counting & Permanence & Reference & Imitation & All tasks \\
  \midrule
  $\pi_{0.5}$ \citep{pi052025} & 28.8 & 17.0 & 17.2 & 8.8 & 17.9 \\
  $\pi_{0.5}$ with past actions & 29.1 & 22.8 & 15.9 & 11.2 & 19.7 \\
  SAM2Act+ \citep{sam2act2025} & 35.3 & 26.0 & 16.8 & 7.3 & 21.4 \\
  SimpleSG + Qwen3-VL-4B & 44.6 & 19.6 & 25.2 & 26.6 & 29.0 \\
  GroundSG + Qwen3-VL-4B & 38.0 & 39.3 & 31.6 & 21.9 & 32.7 \\
  MemER \citep{memer2025} & 48.8 & \textbf{53.2} & \textbf{38.0} & 29.5 & 42.4 \\
  FrameSamp + Expert & 66.8 & 25.2 & 24.1 & 28.9 & 36.3 \\
  FrameSamp + Modulator & 65.2 & 25.1 & 36.3 & \textbf{51.4} & 44.5 \\
  WeaveLA \citep{weavela2026} & 31.0 & 19.5 & 12.5 & 9.5 & 18.1 \\
  $\pi_{0.5}$ + AGM \citep{agm2026} & 56.0 & -- & -- & -- & -- \\
  \addlinespace[2pt]
  \ours No Reasoning & 79.2\,{\scriptsize$\pm$\,1.0} & 23.8\,{\scriptsize$\pm$\,2.6} & 35.2\,{\scriptsize$\pm$\,2.0} & 48.0\,{\scriptsize$\pm$\,1.0} & 46.5\,{\scriptsize$\pm$\,0.9} \\
  \ours Explicit Reasoning & 81.5\,{\scriptsize$\pm$\,1.8} & 26.8\,{\scriptsize$\pm$\,0.6} & 27.7\,{\scriptsize$\pm$\,2.0} & 47.0\,{\scriptsize$\pm$\,2.6} & 45.8\,{\scriptsize$\pm$\,1.5} \\
  \ours \ourmethod{} & \textbf{82.7}\,{\scriptsize$\pm$\,3.8} & 23.8\,{\scriptsize$\pm$\,3.1} & 37.8\,{\scriptsize$\pm$\,2.9} & 48.0\,{\scriptsize$\pm$\,1.5} & \textbf{48.1}\,{\scriptsize$\pm$\,1.0} \\
  \bottomrule
 \end{tabular}
\end{table}
\lsn Trained on all 16 tasks (Table~\ref{tab:memory}), No Reasoning, Explicit Reasoning, and \ourmethodit{} score 46.5, 45.8, and 48.1, respectively, compared with 44.5 for FrameSamp + Modulator \citep{robomme2026}, the benchmark's best published result. On the counting tasks, \ourmethodit{}'s 82.7 is within 1.2 points of the oracle-subgoal rows (83.9 and 82.6; \citealp{robomme2026}), which receive the simulator's ground-truth subgoals, and above the trained symbolic systems (38.0 to 48.8), which run a second vision-language model to generate their subgoals. Additionally, we sweep the cross-entropy weight over three values in both training setups (Table~\ref{tab:robommegrid}). \ourmethodit{} scores above No Reasoning in all six settings and is the best of the three policies in five, within 1.2 points of Explicit Reasoning in the sixth.

\paragraph{RMBench.} \lsn On RMBench (Table~\ref{tab:rmbench}), the three policies are within a point of one another on Blocks Ranking Try, Battery Try, and Cover Blocks. On Press Button, whose count exists only in the ledger, Explicit Reasoning and \ourmethodit{} score 73.3 and 74.3, respectively, compared with 7.7 for No Reasoning. Both keep a correct ledger in 99.7\% of episodes (Table~\ref{tab:ledger}). The four-task means are 60.4, 77.0, and 77.5 for No Reasoning, Explicit Reasoning, and \ourmethodit{}, respectively. All three are above the four-task mean of WLA-0 (56.5; \citealp{wla2026}).

\begin{table}[t]
 \centering
 \caption{\lsn \textbf{RMBench.} Success (\%) on each task (100 cases) and the four-task mean. Shaded rows are ours, mean $\pm$ standard deviation over three evaluation seeds. Published rows are from the benchmark paper \citep{rmbench2026}, EventVLA \citep{eventvla2026}, and the systems' own papers (Appendix~\ref{app:protocol}).}
 \label{tab:rmbench}
 \small
 \setlength{\tabcolsep}{3pt}
 \begin{tabular}{@{}L{0.42\linewidth}ccccc@{}}
  \toprule
  Policy & \shortstack[c]{Blocks\\Ranking Try} & \shortstack[c]{Battery\\Try} & \shortstack[c]{Press\\Button} & \shortstack[c]{Cover\\Blocks} & \shortstack[c]{Four-task\\mean} \\
  \midrule
  DP \citep{diffusionpolicy2023} & 10 & 10 & 0 & 0 & 5.0 \\
  ACT \citep{zhao2023act} & 0 & 19 & 0 & 0 & 4.8 \\
  $\pi_{0.5}$ \citep{pi052025} & 6 & 16 & 0 & 0 & 5.5 \\
  X-VLA \citep{xvla2025} & 1 & 26 & 0 & 2 & 7.3 \\
  WLA-0 \citep{wla2026} & 23 & \textbf{45} & 74 & 84 & 56.5 \\
  HiMem-WAM \citep{himemwam2026} & 24 & 28 & 8 & 19 & 19.8 \\
  QwenOFT & 37 & 14 & 0 & 0 & 12.8 \\
  MemER \citep{memer2025} & 0 & 27 & 0 & 6 & 8.3 \\
  MemoryVLA (QwenOFT; \citealp{memoryvla2026}) & 53 & 33 & 0 & 69 & 38.8 \\
  EventVLA \citep{eventvla2026} & 81 & 35 & 3 & 97 & 54.0 \\
  \ours No Reasoning & 94.3\,{\scriptsize$\pm$\,2.5} & 42.0\,{\scriptsize$\pm$\,1.7} & 7.7\,{\scriptsize$\pm$\,2.1} & 97.7\,{\scriptsize$\pm$\,2.1} & 60.4\,{\scriptsize$\pm$\,1.5} \\
  \ours Explicit Reasoning & 94.3\,{\scriptsize$\pm$\,1.2} & 42.3\,{\scriptsize$\pm$\,1.5} & 73.3\,{\scriptsize$\pm$\,2.5} & \textbf{98.0}\,{\scriptsize$\pm$\,1.0} & 77.0\,{\scriptsize$\pm$\,1.9} \\
  \ours \ourmethod{} & \textbf{94.7}\,{\scriptsize$\pm$\,1.5} & 43.0\,{\scriptsize$\pm$\,2.6} & \textbf{74.3}\,{\scriptsize$\pm$\,2.5} & \textbf{98.0}\,{\scriptsize$\pm$\,1.0} & \textbf{77.5}\,{\scriptsize$\pm$\,1.1} \\
  \bottomrule
 \end{tabular}
\end{table}

\begin{table}[t]
 \vspace{0.9\baselineskip}
 \sbox{\liberobox}{%
  \small
  \setlength{\tabcolsep}{5pt}%
  \begin{tabular}{@{}lcc@{}}
   \toprule
   Policy & Plus & Standard \\
   \midrule
   DreamVLA \citep{dreamvla2025} & 48.9 & 92.6 \\
   MolmoAct$^\dagger$ \citep{molmoact2025} & 60.8 & 86.6 \\
   VLA-Adapter \citep{vlaadapter2025} & 59.1 & 97.3 \\
   MergeVLA \citep{mergevla2025} & 72.4 & 96.7 \\
   LoopOFT$_\alpha$ \citep{loopvla2026} & 65.8 & 95.3 \\
   Evo-Depth \citep{evodepth2026} & 69.6 & 95.4 \\
   WLA-0 \citep{wla2026} & -- & \textbf{98.6} \\
   $\pi_{0.5}$ \citep{pi052025} & 69.0 & 87.8 \\
   MolmoAct2 \citep{molmoact22026} & 63.6 & 93.5 \\
   \ours No Reasoning & 67.6\,{\scriptsize$\pm$\,0.4} & 98.5\,{\scriptsize$\pm$\,0.3} \\
   \ours Explicit Reasoning & \textbf{73.5}\,{\scriptsize$\pm$\,0.3} & 98.1\,{\scriptsize$\pm$\,0.4} \\
   \ours \ourmethod{} & 72.9\,{\scriptsize$\pm$\,0.3} & \textbf{98.6}\,{\scriptsize$\pm$\,0.3} \\
   \bottomrule
  \end{tabular}}%
 \setlength{\abovecaptionskip}{0pt}%
 \sbox{\liberocap}{\parbox[t]{\textwidth}{\caption{\lsn \hangindent=-\dimexpr\wd\liberobox+1em\relax \hangafter=0 \textbf{LIBERO and LIBERO-Plus.} Success (\%) on LIBERO-Plus (10{,}030 cases) and the four standard suites (2{,}000 cases; per suite in Table~\ref{tab:liberofull}). Shaded rows are ours, mean~$\pm$~standard deviation over three evaluation seeds. Published rows are zero-shot LIBERO-Plus results of policies trained, like ours, on the LIBERO demonstrations alone, except the robot-action-pretrained \mbox{MolmoAct} ($\dagger$). The $\pi_{0.5}$ and MolmoAct2 rows are LIT's baselines \citep{lit2026}, trained this way from the vision-language backbones. The LIBERO-Plus values of DreamVLA, MolmoAct, and VLA-Adapter are evaluations by \citet{worldpilot2026}, \citet{anchoralign2026}, and \citet{elan4d2026}, respectively.}\label{tab:libero}}}%
 \noindent\usebox{\liberocap}\llap{\raisebox{-\dimexpr\ht\liberobox+3.6pt\relax}{\usebox{\liberobox}}}%
\end{table}

\paragraph{LIBERO.}\label{sec:ephemeral} \lsn On LIBERO-Plus, reasoning supervision lifts success from 67.6 for No Reasoning to 73.5 for Explicit Reasoning (Table~\ref{tab:libero}), and \ourmethodit{} scores 72.9, within a point of Explicit Reasoning and 5.3 points above No Reasoning. On the four standard suites, which are close to saturation, No Reasoning, Explicit Reasoning, and \ourmethodit{} score 98.5, 98.1, and 98.6, respectively, within a point of one another and of WLA-0 (98.6; \citealp{wla2026}). By perturbation axis, the gain over No Reasoning is largest on changed robot initial states, 11.4 points for Explicit Reasoning and 10.2 for \ourmethodit{} (Table~\ref{tab:plusaxes}).

\lsn Compared with published policies trained, like ours, on the LIBERO demonstrations alone (Table~\ref{tab:libero}), the two reasoning policies hold the highest LIBERO-Plus totals. Policies of our backbone class trained without reasoning supervision score close to No Reasoning (67.6): Qwen3OFT, LoopVLA's baseline of Qwen3-VL-2B with a direct action MLP, scores 68.8 \citep{loopvla2026}, and LIT's $\pi_{0.5}$ baseline, trained from the vision-language backbone, scores 69.0 \citep{lit2026}. No Reasoning is thus a competent policy of its class, and reasoning supervision is what lifts it: \ourmethodit{} generates no text here, so the lift reaches its action through the representation alone (Section~\ref{sec:policies}).

\paragraph{PhysicalAI-AV.}\label{sec:driving} \lsn The three policies are trained on the driving data and evaluated on scenes disjoint from it. In open loop, both reasoning policies lower the displacement error below No Reasoning's 1.166\,m, to 1.140\,m for Explicit Reasoning and 1.116\,m for \ourmethodit{} (Table~\ref{tab:driving}). In closed loop, \ourmethodit{} reaches a scene score of 52.5\% and Explicit Reasoning 52.3\%, compared with 46.4\% for No Reasoning. Both reasoning policies pass more scenes and cause fewer at-fault collisions. On the organizers' published reference rollouts for the same scenes and protocol \citep{alpasimchallenge2026}, Alpamayo-R1 \citep{alpamayo2025}, a 10B model with four cameras, scores 55.0\%. VaVAM \citep{vavam2025} scores 11.0\% under the competition controller and 31.4\% under a linear controller that is not available in the competition, and five unattributed submissions score 5.8\% to 16.8\%. \ourmethodit{} comes within 2.5 points of Alpamayo-R1 with fewer than half its parameters and half its cameras. Appendix~\ref{app:examples} shows, for every benchmark, a successful and a failed episode of both reasoning policies with the reasoning text they generate: in these examples, no failure mode is clearly specific to \ourmethodit{}.

\begin{table}[htbp]
 \centering
 \small
 \caption{\lsn \textbf{PhysicalAI-AV.} Open loop: minADE (m), the minimum average displacement error over six samples on the 1{,}434 held-out segments of the public PhysicalAI-AV release, computed from its released videos, mean $\pm$ standard deviation over evaluation seeds. Closed loop: AlpaSim's metrics on the 441-scene public split of its challenge with one sampled trajectory per call, mean $\pm$ standard deviation over three rollouts per scene. Infractions are at-fault collisions and off-road events. Shaded rows are ours. VaVAM \citep{vavam2025}: the organizers' reference runs \citep{alpasimchallenge2026} under the competition's nonlinear controller, which our policies also use, and under a linear controller that is not available in the competition.}
 \label{tab:driving}
 \setlength{\tabcolsep}{2pt}
 \begin{tabular}{@{}lccccccc@{}}
  \toprule
  Policy & minADE $\downarrow$ & \shortstack[c]{Pass\\rate (\%) $\uparrow$} & \shortstack[c]{Collision\\any (\%) $\downarrow$} & \shortstack[c]{At-fault\\collision (\%) $\downarrow$} & \shortstack[c]{Off-road\\(\%) $\downarrow$} & \shortstack[c]{km per\\infraction $\uparrow$} & \shortstack[c]{Scene\\score (\%) $\uparrow$} \\
  \midrule
  VaVAM (nonlinear) & -- & 15.0\,{\scriptsize$\pm$\,0.7} & 24.3\,{\scriptsize$\pm$\,0.6} & 18.0\,{\scriptsize$\pm$\,0.6} & 28.0\,{\scriptsize$\pm$\,1.1} & 0.14\,{\scriptsize$\pm$\,0.00} & 11.0\,{\scriptsize$\pm$\,0.7} \\
  VaVAM (linear) & -- & 42.1\,{\scriptsize$\pm$\,0.5} & 33.3\,{\scriptsize$\pm$\,0.3} & 16.4\,{\scriptsize$\pm$\,0.7} & 15.2\,{\scriptsize$\pm$\,0.8} & 0.35\,{\scriptsize$\pm$\,0.00} & 31.4\,{\scriptsize$\pm$\,0.6} \\
  \ours No Reasoning & 1.166\,{\scriptsize$\pm$\,0.010} & 50.1\,{\scriptsize$\pm$\,2.2} & 16.1\,{\scriptsize$\pm$\,1.3} & 6.9\,{\scriptsize$\pm$\,1.6} & 9.5\,{\scriptsize$\pm$\,0.5} & 0.87\,{\scriptsize$\pm$\,0.08} & 46.4\,{\scriptsize$\pm$\,2.2} \\
  \ours Explicit Reasoning & 1.140\,{\scriptsize$\pm$\,0.015} & \textbf{56.4}\,{\scriptsize$\pm$\,1.5} & 16.9\,{\scriptsize$\pm$\,0.7} & \textbf{5.5}\,{\scriptsize$\pm$\,0.5} & \textbf{9.0}\,{\scriptsize$\pm$\,0.1} & \textbf{1.01}\,{\scriptsize$\pm$\,0.03} & 52.3\,{\scriptsize$\pm$\,1.4} \\
  \ours \ourmethod{} & \textbf{1.116}\,{\scriptsize$\pm$\,0.011} & 55.8\,{\scriptsize$\pm$\,0.0} & \textbf{15.6}\,{\scriptsize$\pm$\,0.2} & 6.7\,{\scriptsize$\pm$\,0.3} & 10.0\,{\scriptsize$\pm$\,0.9} & 0.87\,{\scriptsize$\pm$\,0.08} & \textbf{52.5}\,{\scriptsize$\pm$\,0.3} \\
  \bottomrule
 \end{tabular}
\end{table}

\subsection{Latency}
\label{sec:latency}

\begin{figure}[t]
 \centering
 \includegraphics[width=\linewidth]{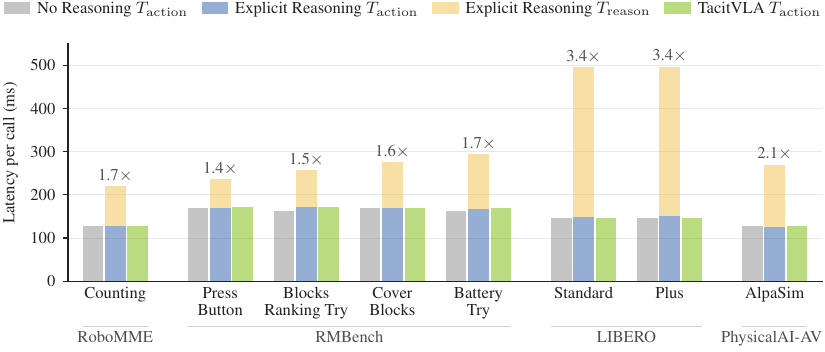}
 \caption{\lsn \textbf{\ourmethodc{} matches the latency of \noreasonc{} on every benchmark, and \explicitc{}'s latency is 1.4$\times$ to 3.4$\times$ higher.} Each bar is one policy's latency per call: $\Tact$ for No Reasoning and \ourmethod{}, and $\Tact + \Treason$ for Explicit Reasoning, whose actions wait for text generation. The ratio above each group is Explicit Reasoning's latency over \ourmethod{}'s. On the memory tasks, \ourmethod{} generates the ledger entry during execution, taking at most 23\% of the execution window, so the next call never waits for generation. All values are means over model calls on one H100, with the per-stage breakdown in Table~\ref{tab:stages} (Appendix~\ref{app:timing}).}
 \label{fig:latency}
\end{figure}

\lsn We time every model stage with per-call timers during evaluation episodes (Appendix~\ref{app:timing}). The three policies differ only in the stages they execute (Figure~\ref{fig:latency}, Table~\ref{tab:stages}). On the memory tasks, the deferred generation finished within the execution window in every call, so generation adds nothing to \ourmethodit{}'s latency (Equation~\ref{eq:cost}). The success rates of Section~\ref{sec:performance} hold unchanged under this schedule: \ourmethodit{}'s action never conditions on $r_t$, and the entry enters the ledger at the same boundary under both policies. Deferring generation therefore changes only when the entry is generated.

\lsn On the RoboMME counting tasks, the action path takes $\Tact=127$\,ms and generating the ledger entry takes $\Treason=93$\,ms under both reasoning policies. \ourmethodit{} generates the entry during execution, in 12\% of the 800\,ms window (16 actions at 20\,Hz), so its action is ready after 127\,ms. Explicit Reasoning runs the two stages in series, and its action is ready after 220\,ms, 1.7$\times$ later. On RMBench, generating the next subtask entry takes 66 to 120\,ms, at most 1.1\% of the 7.0 to 11.2\,s window in which the motion planner executes each plan (Appendix~\ref{app:timing}). \ourmethodit{}'s action is ready after 169 to 171\,ms and Explicit Reasoning's after 236 to 294\,ms, 1.4$\times$ to 1.7$\times$ later.

\lsn On the non-memory tasks, \ourmethodit{} generates no text at deployment (Section~\ref{sec:defer}), so its latency is No Reasoning's: 147\,ms per call on LIBERO and LIBERO-Plus and 127\,ms in closed-loop driving. Explicit Reasoning's action is ready after 495\,ms on LIBERO and 269\,ms in driving, 3.4$\times$ and 2.1$\times$ later, respectively. The difference is the generation of reasoning text the next call never uses. LIBERO's window is 250\,ms (5 actions at 20\,Hz), which \ourmethodit{} meets and Explicit Reasoning does not.

\section{Conclusion}
\label{sec:conclusion}

\lsn On the four benchmarks we study, spanning manipulation and autonomous driving, memory and non-memory tasks, an action expert that never conditions on the current reasoning matches one that does: the benefit of reasoning reaches the action through the reasoning supervision and, on memory tasks, through the ledger of earlier entries in the action expert's context. \ourmethodit{} thus delivers Explicit Reasoning's success at No Reasoning's latency. On the memory tasks, it generates the ledger entry within the execution window at every call. The design remains one backbone feeding one action expert, with no second model or scheduler added. Any reasoning VLA of this form can therefore be trained as \ourmethodit{} by masking the reasoning tokens from the action expert, and deployed with generation omitted or, when the reasoning text serves as memory, run during action execution.

\paragraph{Limitations and future work.} \lsn As for other reasoning VLAs, \ourmethodit{}'s performance depends on the quality of the reasoning data. On non-memory tasks, it can still generate the reasoning text during action execution (Figure~\ref{fig:method}), which could serve interpretability in safety-critical settings such as autonomous driving, a use we leave to future work. Fast-WAM \citep{fastwam2026} reaches an analogous conclusion for future video: a world-action model that retains video co-training but skips future prediction at test time remains competitive with imagine-then-execute variants. Combining the two suggests a backbone with three experts for reasoning, future video, and action, trained on all three targets, yet acting efficiently without first generating reasoning or future video. On memory tasks, the deferred entry adds no latency only while it fits in the execution window. It takes at most 23\% of the window in our benchmarks, and longer entries or shorter windows would add the time by which generation exceeds the window (Equation~\ref{eq:cost}).

\section*{Acknowledgments}

We thank NVIDIA for the compute that made this work possible, and our colleagues in the NVIDIA ASPIRE Group for helpful discussions and feedback.

\setcitestyle{numbers}
\renewcommand{\natexlab}[1]{#1}
\bibliographystyle{plainnat}
\bibliography{references}
\setcitestyle{authoryear,round}

\clearpage
\appendix

\section{Implementation and evaluation protocol}
\label{app:protocol}

\paragraph{Initialization and architecture.} \lsn RoboMME initializes the backbone and action expert from the $\pi_{0.5}$ base weights \citep{pi052025} and adds RoboMME's frame-sampling memory \citep{robomme2026}. RMBench and LIBERO initialize the backbone from the released RynnBrain-2B \citep{rynnbrain2026}, and driving from Qwen3-VL-2B-Instruct \citep{qwen3vl2025}, enlarged with trajectory tokens, each with a randomly initialized action expert. Table~\ref{tab:params} counts the parameters of each policy and Table~\ref{tab:arch} lists the architectures. On every benchmark, the action expert is a second transformer stack with the backbone's attention geometry and no vocabulary: at layer $\ell$, its positions attend to the backbone's layer-$\ell$ keys and values together with their own, so the two stacks form one attention over one sequence while keeping separate weights (Section~\ref{sec:setup}). The action positions continue the prefix's position IDs and attend to one another without a causal mask. Each action position is one action of the chunk. On RoboMME, the flow time enters through adaptive RMSNorm as in $\pi_{0.5}$, and the frame-sampling memory adds a memory attention over sampled frames and a memory-conditioned modulation in every layer of both stacks. On RMBench and LIBERO, the action expert embeds the noised action with the flow time and appends one clean state token. The driving action expert encodes each waypoint with Fourier features and outputs acceleration and curvature, integrated by a unicycle model.

\begin{table}[h]
 \centering
 \small
 \caption{\lsn \textbf{Parameters of the policies (billions).} The tied output head is counted once. The components sum to Explicit Reasoning's total. On RoboMME, the total also includes the 0.315B frame-sampling memory, whose history frames pass through the same frozen SigLIP as the current image. A policy that generates no text (No Reasoning, and \ourmethod{} on the non-memory tasks) uses only the keys and values of the last language layer and not the rest of that layer or the final norm.}
 \label{tab:params}
 \setlength{\tabcolsep}{4pt}
 \begin{tabular}{@{}lrrrrrrr@{}}
  \toprule
  & \multicolumn{4}{c}{Components} & \multicolumn{3}{c}{Policy} \\
  \cmidrule(lr){2-5} \cmidrule(l){6-8}
  Benchmark & \shortstack[r]{Vision\\tower} & \shortstack[r]{Text\\embedding} & \shortstack[r]{Language\\layers} & \shortstack[r]{Action\\expert} & \shortstack[r]{No\\Reasoning} & \shortstack[r]{Explicit\\Reasoning} & \ourmethod{} \\
  \midrule
  RoboMME & 0.415 & 0.527 & 1.982 & 0.430 & 3.547 & 3.669 & 3.669 \\
  RMBench & 0.407 & 0.311 & 1.409 & 0.446 & 2.527 & 2.573 & 2.573 \\
  LIBERO & 0.407 & 0.311 & 1.409 & 0.446 & 2.527 & 2.573 & 2.527 \\
  PhysicalAI-AV & 0.407 & 0.313 & 1.409 & 1.774 & 3.857 & 3.903 & 3.857 \\
  \bottomrule
 \end{tabular}
\end{table}

\begin{table}[h]
 \centering
 \small
 \caption{\lsn \textbf{Architecture of the policies.} In the output head row, $d$ is the action dimension (14 on RMBench, 7 on LIBERO). The RoboMME policies use the $\pi_{0.5}$ architecture with RoboMME's frame-sampling memory. The RMBench, LIBERO, and driving backbones share the Qwen3-VL-2B architecture and differ in vocabulary.}
 \label{tab:arch}
  \setlength{\tabcolsep}{4pt}
 \begin{tabular}{@{}L{0.125\linewidth}L{0.165\linewidth}L{0.205\linewidth}L{0.205\linewidth}L{0.205\linewidth}@{}}
  \toprule
  Component & Setting & RoboMME & RMBench, LIBERO & PhysicalAI-AV \\
  \midrule
  \multirow{2.4}{=}{Vision encoder} & layers & 27 (SigLIP) & 24 & 24 \\ \arrayrulecolor{black!22}\cmidrule(l){2-5}\arrayrulecolor{black}
   & width & 1{,}152 & 1{,}024 & 1{,}024 \\
  \midrule
  \multirow{5.2}{=}{Language model} & layers & 18 (Gemma-2B) & 28 & 28 \\ \arrayrulecolor{black!22}\cmidrule(l){2-5}\arrayrulecolor{black}
   & width & 2{,}048 & 2{,}048 & 2{,}048 \\ \arrayrulecolor{black!22}\cmidrule(l){2-5}\arrayrulecolor{black}
   & heads (query / key-value) & 8 / 1 & 16 / 8 & 16 / 8 \\ \arrayrulecolor{black!22}\cmidrule(l){2-5}\arrayrulecolor{black}
   & feed-forward & 16{,}384, GeGLU & 6{,}144, SwiGLU & 6{,}144, SwiGLU \\
  \midrule
  \multirow{11}{=}{Action expert} & layers & 18 & 28 & 28 \\ \arrayrulecolor{black!22}\cmidrule(l){2-5}\arrayrulecolor{black}
   & width & 1{,}024 & 1{,}024 & 2{,}048 \\ \arrayrulecolor{black!22}\cmidrule(l){2-5}\arrayrulecolor{black}
   & heads (query / key-value) & 8 / 1 & 16 / 8 & 16 / 8 \\ \arrayrulecolor{black!22}\cmidrule(l){2-5}\arrayrulecolor{black}
   & feed-forward & 4{,}096, GeGLU & 3{,}072 & 8{,}256 \\ \arrayrulecolor{black!22}\cmidrule(l){2-5}\arrayrulecolor{black}
   & coupling & shared attention with Gemma, layer by layer & attends to backbone keys and values, layer by layer & as RMBench and LIBERO \\ \arrayrulecolor{black!22}\cmidrule(l){2-5}\arrayrulecolor{black}
   & time conditioning & adaptive RMSNorm (scale, shift, gate) & sinusoidal, fused into each action & Fourier features per waypoint \\
  \midrule
  \multirow{9.2}{=}{Action interface} & actions per chunk & 20 & 32 (RMBench), 10 (LIBERO) & 64 \\ \arrayrulecolor{black!22}\cmidrule(l){2-5}\arrayrulecolor{black}
   & action dimension & 32 (8 active) & 14, 7 & 2: acceleration, curvature \\ \arrayrulecolor{black!22}\cmidrule(l){2-5}\arrayrulecolor{black}
   & state & as $\pi_{0.5}$ & one token, 14-D or 8-D & 16 past ego states in the prefix \\ \arrayrulecolor{black!22}\cmidrule(l){2-5}\arrayrulecolor{black}
   & output head & linear $1{,}024 {\to} 32$ & MLP $1{,}024 {\to} 1{,}024 {\to} d$ & linear $2{,}048 {\to} 2$ \\
  \midrule
  Flow matching & sampler, solver, steps & beta, Euler, 10 & beta, Euler, 32 & beta, Euler, 10 \\
  \bottomrule
 \end{tabular}
\end{table}

\paragraph{Training data.} \lsn RoboMME counting uses the 400 demonstrations of BinFill, PickXtimes, SwingXtimes, and StopCube, and the 16-task policy uses all 1{,}600, with subtask text on the four counting tasks only. On RMBench, each task has its own policy, trained on the benchmark's 50 released demonstrations. LIBERO trains on the 1{,}817 demonstrations that replay successfully in the simulator, of which 1{,}616 carry DeepThinkVLA reasoning annotations \citep{deepthink2025}. Driving trains on 3.8 million reasoning-labeled events, each with a 1.6\,s history and a 6.4\,s future window.

\begin{table}[h]
 \centering
 \caption{\lsn \textbf{Training settings.} Learning rates are for the vision tower, the language model, and the action expert, respectively, and RoboMME uses one rate for all three.}
 \label{tab:training}
 \setlength{\tabcolsep}{5pt}
 \begin{tabular}{@{}lrlrrr@{}}
  \toprule
  Setting & Batch & Peak learning rate & Warmup & Updates & Noise draws \\
  \midrule
  RoboMME counting & 64 & $5{\times}10^{-5}$ & 5k & 60k & 1 \\
  RoboMME all tasks & 64 & $5{\times}10^{-5}$ & 10k & 80k & 1 \\
  RMBench & 512 & $5{\times}10^{-6}$ / $5{\times}10^{-5}$ / $2{\times}10^{-4}$ & 2k & 80k & 16 \\
  LIBERO & 512 & $5{\times}10^{-6}$ / $5{\times}10^{-5}$ / $2{\times}10^{-4}$ & 2k & 40k & 16 \\
  PhysicalAI-AV & 256 & $1{\times}10^{-5}$ / $1{\times}10^{-4}$ / $4{\times}10^{-4}$ & 4k & 100k & 1 \\
  \bottomrule
 \end{tabular}
\end{table}

\paragraph{Optimization.} \lsn Table~\ref{tab:training} lists batch sizes, learning rates, and schedules. All recipes use AdamW. RMBench and LIBERO decay the learning rate by cosine to 10\% of its peak, driving to 1\%, and RoboMME holds it constant after warmup. The cross-entropy weight $\lambda$ is 0.1 on LIBERO and driving and 1.0 on RMBench, and on RoboMME each policy is trained at $\lambda \in \{0.1, 0.01, 0.001\}$ and reported at its best (Table~\ref{tab:robommegrid}). The same $\lambda$ scales the tokenized-action term of Equation~\ref{eq:loss} in every policy, and in No Reasoning it scales that term alone. The gradient from the action expert into the backbone is enabled after 2k updates on RMBench and LIBERO, after the 4k warmup in driving, and from the start on RoboMME. \ourmethodit{}'s updates cost slightly less than Explicit Reasoning's, whose action positions also attend to $r_t$, and No Reasoning's sequences carry no reasoning text, so its updates are the cheapest.

\paragraph{Context.} \lsn On RoboMME, the context $c_t$ is the front and wrist views, the benchmark's frame-sampling memory of up to 32 history frames, the robot state, the instruction, and, on the counting tasks, the ledger. On RMBench, it is, in order, the instruction, the episode's first head frame, the ledger of prior subtasks, the head frame from 32 actions earlier, the current head, left, and right frames, the robot state, and the cue for the next subtask. On LIBERO, it is the agent and wrist images, the agent image from eight steps earlier, the robot state, and the instruction. In driving, it is four frames from each of two front cameras at 10\,Hz, 16 past ego states, and a fixed reasoning prompt. No Reasoning keeps the same observation on every benchmark, including RMBench's first and lagged frames, and drops only the ledger.

\paragraph{Prompts.} \lsn Reasoning examples and direct examples use separate prompts, and the reasoning prompt adds a fixed cue that requests the reasoning. On LIBERO, the annotated demonstrations train under the reasoning prompt and the 201 unannotated ones under the direct prompt, for action supervision only. RoboMME and RMBench examples carry the ledger prompt, and driving examples the reasoning prompt. Both reasoning policies are deployed with the prompt they trained on, including \ourmethodit{} on the non-memory tasks, where it generates no text.

\paragraph{Reasoning targets.}\label{app:targets} \lsn On LIBERO, for the instruction ``turn on the stove and put the moka pot on it'', DeepThinkVLA's released per-frame annotation at the first frame reads:
\begin{quote}\small
This is the initial frame. The stove (right-front) is off. The moka pot (center-front) and a pan (left-front) are on the counter. The stove knob (right-front, on stove) is accessible. To begin the global task, the stove must be turned on first. Turn on the stove.
\end{quote}
The prompt is ``Task: \textit{instruction}'', followed on reasoning examples by ``Current reasoning:''. On RMBench, each call's prompt carries the ledger so far under ``Previous subtask:'' and asks for ``Current subtask:''. In driving, one training annotation reads, verbatim: ``Maintain lane and search for a usable gap to the left to prepare for a lane change because a car is overtaking alongside in the left lane while a slower SUV is ahead in the current lane.'' The prompt asks the model to ``output the chain-of-thought reasoning of the driving process, then output the future trajectory''.

\begin{table}[h]
 \centering
 \small
 \caption{\lsn \textbf{Observation, action, and execution.} Solver steps are uniform Euler steps. Closed-loop driving replans at every 100\,ms control step of the simulator, so one waypoint is executed per call.}
 \label{tab:actionprotocol}
 \setlength{\tabcolsep}{4pt}
 \begin{tabular}{@{}lL{0.37\linewidth}L{0.22\linewidth}rr@{}}
  \toprule
  Benchmark & Observation & Predicted chunk & Executed & \shortstack[r]{Solver\\steps} \\
  \midrule
  RoboMME & front and wrist views, up to 32 sampled history frames & 20 actions (32-D, 8 active) & 16 & 10 \\ \arrayrulecolor{black!22}\cmidrule{1-5}\arrayrulecolor{black}
  RMBench & first frame, head frame 32 actions earlier, current head/left/right & 32 targets (14-D) & 32 & 32 \\ \arrayrulecolor{black!22}\cmidrule{1-5}\arrayrulecolor{black}
  LIBERO & agent and wrist images, agent image 8 steps earlier & 10 actions (7-D) & 5 & 32 \\ \arrayrulecolor{black!22}\cmidrule{1-5}\arrayrulecolor{black}
  PhysicalAI-AV & 4 frames from two front cameras, 16 past ego states & 64 waypoints / 6.4\,s & 1 & 10 \\
  \bottomrule
 \end{tabular}
\end{table}

\paragraph{Evaluation protocol.} \lsn Table~\ref{tab:actionprotocol} lists observations, action chunks, and solver steps. RoboMME is evaluated as the benchmark specifies: its official test episodes (50 per task), horizon, chunking, and the three seeds its evaluation script names. RMBench evaluates the benchmark's own 100 cases per task, disjoint from the 50 demonstrations and identical for every policy and evaluation seed. LIBERO evaluates 50 initial states per task (2{,}000 cases) under the DeepThinkVLA step caps, and LIBERO-Plus its 10{,}030 published variants under the same caps, with the perturbations applied as the benchmark's released wrapper applies them. Reasoning is decoded greedily on the three robot benchmarks, up to 128 tokens (RoboMME) or 256 (RMBench, LIBERO), and sampled in driving at temperature 0.6 and top-p 0.98. Open-loop driving reports the minimum over six sampled trajectories of the average displacement error on the 1{,}434 held-out segments, averaged over action-sampling seeds. Closed loop follows the protocol of the AlpaSim E2E Closed Loop Challenge on its 441-scene public split \citep{alpasimchallenge2026}: 20\,s rollouts at 10\,Hz under the challenge's controller and scoring, three rollouts per scene, with a failed rollout counted as a failure. The driver samples one trajectory per call and executes it, the protocol of the challenge's reference driver. The route is not a model input.

\paragraph{Ledger semantics.} \lsn On RoboMME, one subtask sentence, the benchmark's simple subgoal without the image coordinates of its grounded variant, is generated per call. An entry enters the ledger only after its 16-action chunk has executed, so the first entry appears at the second call. Adjacent duplicates are removed, and the ledger resets per episode and is never truncated. Explicit Reasoning generates its entry before the action it conditions on and \ourmethodit{} after computing the action. Both append it at the same boundary, so every later call conditions on a ledger built under one protocol. In both policies, an entry names the subtask being undertaken and carries no outcome: a grasp that fails still leaves its entry, and the count recovers when the next call generates the subtask again. On RMBench, the next subtask is decoded greedily for each 32-action plan.

\paragraph{Deployment.} \lsn On the memory tasks, the deferred generation continues from the action path's key-value cache after the action is computed, and if it has not finished when the next observation arrives, the next call waits for it (Equation~\ref{eq:cost}). In training, the ledger prefix is the annotated subtask sequence, and at deployment it is generated, a shift that Explicit Reasoning faces as well. On the non-memory tasks, \ourmethodit{} can still generate the reasoning text after the action when it is wanted for inspection.

\paragraph{Baselines.} \lsn Published rows follow each benchmark's protocol and are quoted as printed, except means over our task sets that a source does not print, which we compute from its per-task values. RoboMME rows are the RoboMME paper's \citep{robomme2026}, except WeaveLA and AGM, which report their own. RMBench rows include QwenOFT and the MemER and MemoryVLA rows from EventVLA's reproductions \citep{eventvla2026}. LIBERO-Plus rows are zero-shot policies trained on the LIBERO demonstrations alone, plus the robot-action-pretrained MolmoAct for reference; the $\pi_{0.5}$ and MolmoAct2 rows are LIT's baselines \citep{lit2026}, and the LIBERO-Plus values of MolmoAct, DreamVLA, and VLA-Adapter come from \citet{anchoralign2026}, \citet{worldpilot2026}, and \citet{elan4d2026}, respectively. Systems with robot-action pretraining fall outside this matched comparison, such as DeepThinkVLA (79.0; \citealp{deepthink2025}) on LIBERO-Plus, as do systems with significantly more parameters than ours, such as VLA-GSE (81.2; \citealp{vlagse2026}) and FoMoVLA (80.5; \citealp{fomovla2026}) on LIBERO-Plus and MemoryWAM (81.5; \citealp{memorywam2026}) on RMBench. 

\section{Ablations and full result tables}
\label{app:ablations}

\lsn Tables~\ref{tab:robommegrid} to \ref{tab:ledger} hold the RoboMME weight sweep, the per-axis and per-suite LIBERO tables behind Section~\ref{sec:results}, and the ledger correctness counts on the memory tasks.

\begin{table}[htp]
 \centering
 \caption{\lsn \textbf{RoboMME, the cross-entropy weight sweep.} Success (\%) of the three policies at $\lambda \in \{0.1, 0.01, 0.001\}$, mean $\pm$ standard deviation over three evaluation seeds: the Counting-suite policies on the four counting tasks (200 cases) and the 16-task policies on all 16 tasks (800 cases).}
 \label{tab:robommegrid}
 \setlength{\tabcolsep}{6pt}
 \begin{tabular}{@{}lccccccc@{}}
  \toprule
  & \multicolumn{3}{c}{Trained on the Counting suite} & & \multicolumn{3}{c}{Trained on all suites} \\
  \cmidrule{2-4} \cmidrule{6-8}
  Policy & $\lambda{=}0.1$ & $0.01$ & $0.001$ & & $\lambda{=}0.1$ & $0.01$ & $0.001$ \\
  \midrule
  No Reasoning & 77.3\,{\scriptsize$\pm$\,2.5} & 79.0\,{\scriptsize$\pm$\,1.0} & 74.5\,{\scriptsize$\pm$\,2.8} &  & 43.6\,{\scriptsize$\pm$\,0.2} & 46.5\,{\scriptsize$\pm$\,0.9} & 44.0\,{\scriptsize$\pm$\,1.6} \\
  Explicit Reasoning & \textbf{86.7}\,{\scriptsize$\pm$\,2.0} & 84.7\,{\scriptsize$\pm$\,1.6} & 73.0\,{\scriptsize$\pm$\,2.0} &  & 43.5\,{\scriptsize$\pm$\,1.2} & 45.8\,{\scriptsize$\pm$\,1.5} & 43.8\,{\scriptsize$\pm$\,1.3} \\
  \ourmethod{} & 85.5\,{\scriptsize$\pm$\,3.3} & \textbf{89.5}\,{\scriptsize$\pm$\,1.3} & \textbf{82.5}\,{\scriptsize$\pm$\,2.5} &  & \textbf{44.0}\,{\scriptsize$\pm$\,0.9} & \textbf{47.6}\,{\scriptsize$\pm$\,1.4} & \textbf{48.1}\,{\scriptsize$\pm$\,1.0} \\
  \bottomrule
 \end{tabular}
\end{table}

\begin{table}[htp]
 \centering
 \small
 \caption{\lsn \textbf{LIBERO-Plus by perturbation axis.} Success (\%) on each of the seven perturbation axes and on the full 10{,}030-case suite, mean $\pm$ standard deviation over three evaluation seeds.}
 \label{tab:plusaxes}
 \setlength{\tabcolsep}{2pt}
 \begin{tabular}{@{}lcccccccc@{}}
  \toprule
  Policy & Robot init. & Sensor & Camera & Language & Objects & Background & Light & All \\
  Cases & 1{,}550 & 1{,}601 & 1{,}599 & 1{,}537 & 1{,}525 & 1{,}076 & 1{,}142 & 10{,}030 \\
  \midrule
  No Reasoning & 59.7\,{\scriptsize$\pm$\,1.1} & 51.7\,{\scriptsize$\pm$\,1.1} & 41.0\,{\scriptsize$\pm$\,1.1} & 76.6\,{\scriptsize$\pm$\,0.9} & 73.2\,{\scriptsize$\pm$\,1.0} & 92.9\,{\scriptsize$\pm$\,0.7} & 94.0\,{\scriptsize$\pm$\,0.6} & 67.6\,{\scriptsize$\pm$\,0.4} \\
  Explicit Reasoning & \textbf{71.1}\,{\scriptsize$\pm$\,0.8} & 51.1\,{\scriptsize$\pm$\,0.9} & \textbf{51.5}\,{\scriptsize$\pm$\,0.9} & \textbf{83.7}\,{\scriptsize$\pm$\,0.6} & \textbf{78.3}\,{\scriptsize$\pm$\,0.7} & 97.2\,{\scriptsize$\pm$\,0.3} & \textbf{96.8}\,{\scriptsize$\pm$\,0.4} & \textbf{73.5}\,{\scriptsize$\pm$\,0.3} \\
  \ourmethod{} & 69.9\,{\scriptsize$\pm$\,0.8} & \textbf{60.0}\,{\scriptsize$\pm$\,0.8} & 42.2\,{\scriptsize$\pm$\,0.8} & 83.0\,{\scriptsize$\pm$\,0.6} & 77.1\,{\scriptsize$\pm$\,0.7} & \textbf{98.0}\,{\scriptsize$\pm$\,0.3} & 95.4\,{\scriptsize$\pm$\,0.4} & 72.9\,{\scriptsize$\pm$\,0.3} \\
  \bottomrule
 \end{tabular}
\end{table}

\begin{table}[htp]
 \centering
 \caption{\lsn \textbf{LIBERO, every policy and suite.} Success (\%) on the four standard suites (50 trials per task) and on the 10{,}030-case LIBERO-Plus suite, mean $\pm$ standard deviation over three evaluation seeds.}
 \label{tab:liberofull}
 \setlength{\tabcolsep}{2.5pt}
 \begin{tabular}{@{}lcccccc@{}}
  \toprule
  Policy & Spatial & Object & Goal & Long & Avg. & Plus \\
  \midrule
  No Reasoning & \textbf{99.2}\,{\scriptsize$\pm$\,0.4} & 99.6\,{\scriptsize$\pm$\,0.3} & \textbf{98.8}\,{\scriptsize$\pm$\,0.5} & 96.4\,{\scriptsize$\pm$\,0.9} & 98.5\,{\scriptsize$\pm$\,0.3} & 67.6\,{\scriptsize$\pm$\,0.4} \\
  Explicit Reasoning & 98.4\,{\scriptsize$\pm$\,0.7} & 99.8\,{\scriptsize$\pm$\,0.3} & 97.8\,{\scriptsize$\pm$\,0.9} & 96.4\,{\scriptsize$\pm$\,1.1} & 98.1\,{\scriptsize$\pm$\,0.4} & \textbf{73.5}\,{\scriptsize$\pm$\,0.3} \\
  \ourmethod{} & \textbf{99.2}\,{\scriptsize$\pm$\,0.5} & \textbf{100.0}\,{\scriptsize$\pm$\,0.0} & 98.0\,{\scriptsize$\pm$\,0.7} & \textbf{97.2}\,{\scriptsize$\pm$\,0.8} & \textbf{98.6}\,{\scriptsize$\pm$\,0.3} & 72.9\,{\scriptsize$\pm$\,0.3} \\
  \bottomrule
 \end{tabular}
\end{table}

\FloatFlush

\paragraph{Ledger correctness.}\label{app:ledger} \lsn On the memory tasks, later calls rely on the ledger to count what has been done, so deferring generation is free only if the entries stay correct. Table~\ref{tab:ledger} counts the episodes whose final ledger matches the instructed count: the number of pick-and-place, swing, or bin-fill completions on the RoboMME counting tasks (StopCube, which counts the cube's passes rather than the robot's completions, is not in the table), and the presses per button plus one confirm entry on Press Button. \ourmethodit{} and Explicit Reasoning are within two points of each other on every task, and every \ourmethodit{} failure on Press Button ends on the step budget with the physical press count off target while its ledger is correct.

\begin{table}[htbp]
 \centering
 \caption{\lsn \textbf{Correct ledgers (\%).} Episodes whose final generated ledger matches the instructed count, for the Counting-suite policies of Table~\ref{tab:counting} and the Press Button policies of Table~\ref{tab:rmbench}.}
 \label{tab:ledger}
 \setlength{\tabcolsep}{6pt}
 \begin{tabular}{@{}lcccc@{}}
  \toprule
  Policy & BinFill & PickXtimes & SwingXtimes & Press Button \\
  \midrule
  Explicit Reasoning & 90.0 & 100.0 & 100.0 & 99.7 \\
  \ourmethod{} & 92.0 & 100.0 & 100.0 & 99.7 \\
  \bottomrule
 \end{tabular}
\end{table}

\FloatBarrier
\section{Latency accounting}
\label{app:timing}

\begin{table}[h!]
 \centering
 \footnotesize
 \caption{\lsn \textbf{Stage times and the latency per call (ms).} The left columns give the execution window $\Texec$ and each policy's action path $\Tact$ and reasoning cost $\Treason$ (Section~\ref{sec:timing}), and the right columns give the latency per call: $\Tact$ for No Reasoning and \ourmethod{}, and $\Tact+\Treason$ for Explicit Reasoning, whose action waits for the generation. \ourmethod{} generates during $\Texec$ on the memory tasks and not at all on the non-memory tasks. All values are means of per-call stage timers on one H100. RMBench's window is the motion planner's mean time to execute a plan.}
 \label{tab:stages}
 \setlength{\tabcolsep}{2.4pt}
 \footnotesize
 \begin{tabular}{@{}>{\scriptsize}lrrrrrrrrr@{}}
  \toprule
  & & \multicolumn{3}{c}{$\Tact$} & \multicolumn{2}{c}{$\Treason$} & \multicolumn{3}{c}{Latency} \\
  \cmidrule(lr){3-5} \cmidrule(lr){6-7} \cmidrule(lr){8-10}
  \multicolumn{1}{l}{Benchmark} & \multicolumn{1}{r}{\scriptsize $\Texec$} & \multicolumn{1}{r}{\scriptsize \shortstack[c]{No\\Reasoning}} & \multicolumn{1}{r}{\scriptsize \ourmethod{}} & \multicolumn{1}{r}{\scriptsize \shortstack[c]{Explicit\\Reasoning}} & \multicolumn{1}{r}{\scriptsize \ourmethod{}} & \multicolumn{1}{r}{\scriptsize \shortstack[c]{Explicit\\Reasoning}} & \multicolumn{1}{r}{\scriptsize \shortstack[c]{No\\Reasoning}} & \multicolumn{1}{r}{\scriptsize \ourmethod{}} & \multicolumn{1}{r}{\scriptsize \shortstack[c]{Explicit\\Reasoning}} \\
  \midrule
  RoboMME, counting & 800 & 127 & 127 & 127 & 93 & 93 & 127 & 127 & 220 \\
  RMBench, Press Button & 6{,}972 & 170 & 171 & 170 & 66 & 66 & 170 & 171 & 236 \\
  RMBench, Blocks Ranking Try & 8{,}116 & 162 & 171 & 171 & 81 & 87 & 162 & 171 & 257 \\
  RMBench, Cover Blocks & 9{,}359 & 170 & 170 & 169 & 101 & 106 & 170 & 170 & 275 \\
  RMBench, Battery Try & 11{,}181 & 162 & 169 & 168 & 120 & 126 & 162 & 169 & 294 \\
  LIBERO & 250 & 147 & 147 & 149 & -- & 345 & 147 & 147 & 495 \\
  LIBERO-Plus & 250 & 147 & 147 & 150 & -- & 347 & 147 & 147 & 497 \\
  PhysicalAI-AV, closed loop & 100 & 127 & 127 & 126 & -- & 143 & 127 & 127 & 269 \\
  \bottomrule
 \end{tabular}
\end{table}

\paragraph{What is timed.} \lsn Each stage of a call is timed by host wall-clock timers with device synchronization at each stage boundary, on one NVIDIA H100 80\,GB GPU in bf16, one policy process at a time. The visual encoder, the prompt prefill, and each denoising step run as captured CUDA graphs, and text generation replays one graph per token over a static key-value cache. $\Tact$ is visual encoding, prompt prefill, and denoising, the same three stages for every policy. The prompt prefill sits in $\Tact$ for each policy (under Explicit Reasoning it runs as the first step of generation). $\Treason$ is the autoregressive generation of $r_t$ plus, for Explicit Reasoning, the backbone's pass over the generated tokens that its action expert conditions on (5 to 8\,ms), which runs after generation and before denoising. On the memory tasks, \ourmethodit{}'s action and its deferred generation share one prefill, so that prefill is on its action path and its $\Tact$ is a few milliseconds above No Reasoning's.

\paragraph{Execution windows.} \lsn We use the simulators' own control rates. LIBERO executes 5 actions per call at 20\,Hz (250\,ms), RoboMME 16 at 20\,Hz (800\,ms), and closed-loop driving replans at every 100\,ms control step. RMBench has no control rate: an action is a joint-position target, and the simulator hands each of the 32 targets of a plan to a motion planner that moves the arm under its joint velocity and acceleration limits, stepping the physics at 4\,ms until the arm arrives. A plan's duration is set by the commanded motion. We count those physics steps per plan and report the mean over the 100 cases of each task (218 to 349\,ms per target). No deferred generation exceeded its window in any call (the longest RoboMME entry took 180\,ms, the longest RMBench entry 225\,ms), so \ourmethodit{}'s latency is $\Tact$ in every cell.

\paragraph{Reading the table.} \lsn On the non-memory tasks, \ourmethodit{} and No Reasoning run the same stages, and their latencies agree to the millisecond. On the memory tasks, both reasoning policies generate entries of similar cost (RMBench $\Treason$ 66 to 120 compared with 66 to 126\,ms, the latter including the pass over the entry), and they differ only in where that generation sits (Section~\ref{sec:latency}). The RoboMME row is the 16-task policy on its four counting tasks. On its other 12 tasks, which keep no ledger, no policy generates text and the three latencies coincide (143 to 144\,ms).

\section{Qualitative results}
\label{app:examples}

\begin{figure}[!htbp]
 \centering
 \includegraphics[width=\linewidth,height=0.9\textheight,keepaspectratio]{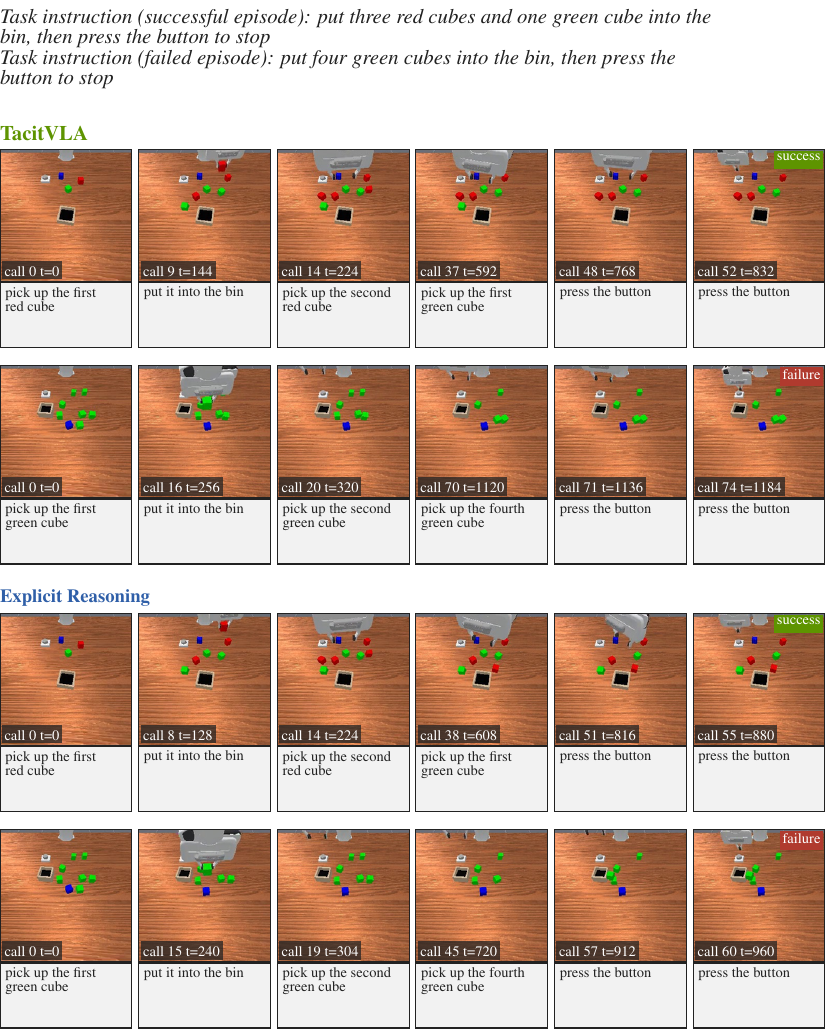}
 \caption{\lsn \textbf{RoboMME BinFill.} The front camera at the calls that generate a new ledger entry and at the episode's last call. Counting-suite policies of Table~\ref{tab:counting}.}
 \label{fig:qual_robomme}
\end{figure}

\begin{figure}[p]
 \centering
 \includegraphics[width=\linewidth,height=0.9\textheight,keepaspectratio]{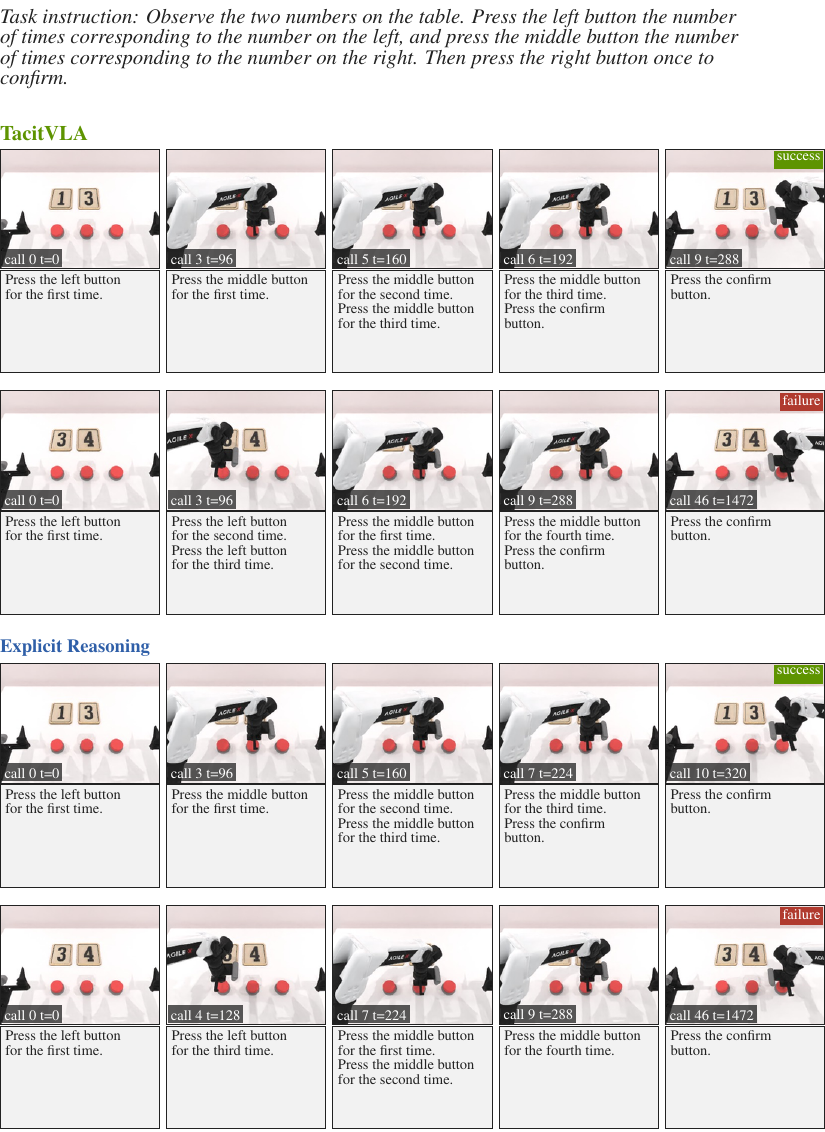}
 \caption{\lsn \textbf{RMBench Press Button.} The head camera at the calls that generate a new ledger entry and at the episode's last call. A call can add two entries when a press completes within the action chunk. Policies of Table~\ref{tab:rmbench}.}
 \label{fig:qual_rmbench}
\end{figure}

\begin{figure}[p]
 \centering
 \includegraphics[width=\linewidth,height=0.9\textheight,keepaspectratio]{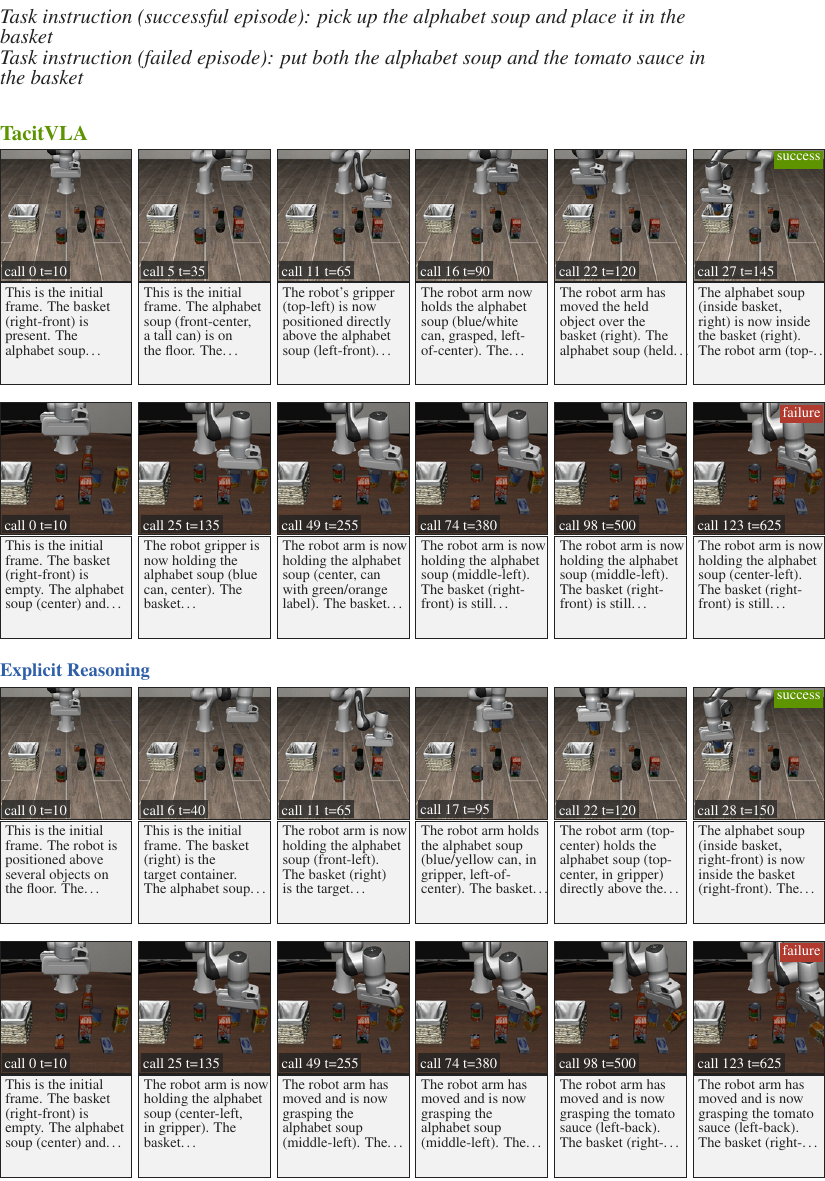}
 \caption{\lsn \textbf{LIBERO.} The agent camera at six calls, with the opening of the reasoning generated at each. Policies of Table~\ref{tab:libero}.}
 \label{fig:qual_libero}
\end{figure}

\begin{figure}[p]
 \centering
 \includegraphics[width=\linewidth,height=0.9\textheight,keepaspectratio]{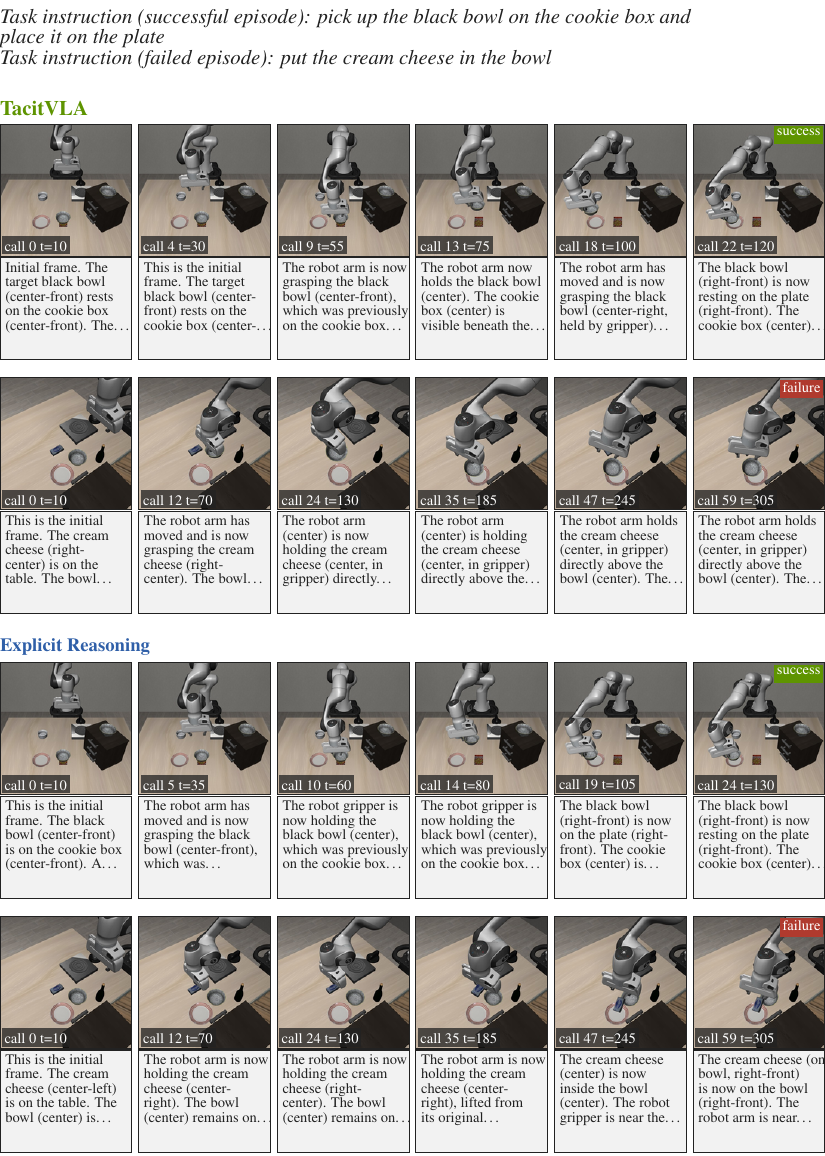}
 \caption{\lsn \textbf{LIBERO-Plus, Camera Viewpoints.} The agent camera on two tasks under perturbed viewpoints. Policies of Table~\ref{tab:libero}.}
 \label{fig:qual_libero_plus}
\end{figure}

\begin{figure}[p]
 \centering
 \includegraphics[width=\linewidth,height=0.9\textheight,keepaspectratio]{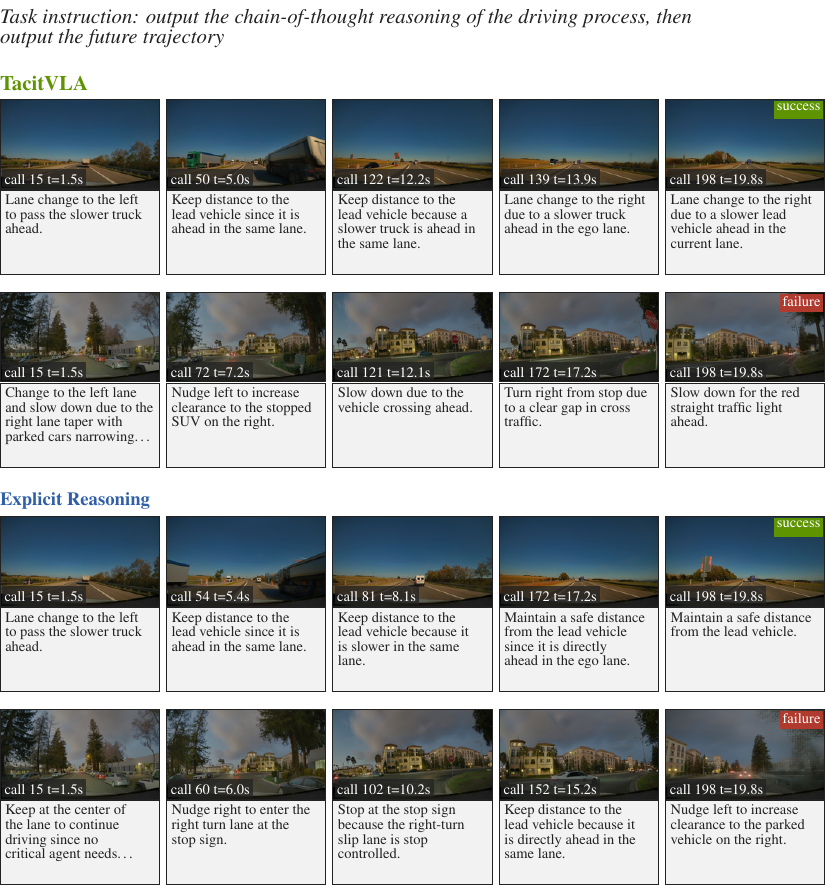}
 \caption{\lsn \textbf{PhysicalAI-AV closed loop.} The front camera at five calls of two public-split scenes, with the chain of causation generated at each call. Policies of Table~\ref{tab:driving}.}
 \label{fig:qual_alpasim}
\end{figure}

\lsn Figures~\ref{fig:qual_robomme} to \ref{fig:qual_alpasim} show the reasoning text the policies generate and the episodes it accompanies, for RoboMME, RMBench, LIBERO, LIBERO-Plus, and driving: a successful episode and a failed one, each rolled out by \ourmethodit{} and by Explicit Reasoning from the same initial state, with the reasoning text generated at each shown call under its frame. The two upper rows are \ourmethodit{} on the successful and then the failed episode, and the two lower rows are Explicit Reasoning on the same two episodes. Frames are stamped with the call index and the simulator step before that call. The box under a frame holds the text the backbone generated at that call, verbatim. On the memory tasks, the text is the ledger entry, and the frames are the calls that generate a new entry, plus the episode's last call; on LIBERO and driving, the text is the reasoning for that call and the frames span the episode. On the memory tasks, the text is generated during execution of the action chunk and enters the ledger for the next call. On LIBERO and driving, where \ourmethodit{} generates no text at deployment, we generated it for this appendix after computing each action, as on the memory tasks, so the actions are unchanged.

\lsn In these examples, no failure mode is clearly specific to \ourmethodit{}. On every failure shown, Explicit Reasoning fails the same episode with the same kind of execution error: cubes that do not end in the bin, a confirm press that never registers, a grasp carried without a placement, a lateral departure from the lane corridor.

\end{document}